%% file: meta-gnn-survey_arxiv.tex
\documentclass{article}
\usepackage{arxiv}

\usepackage{times}
\usepackage{soul}
\usepackage{url}
\usepackage[colorlinks=true, citecolor=blue]{hyperref}
\usepackage[utf8]{inputenc}
\usepackage{caption}
\usepackage{graphicx}
\usepackage{amsmath}
\usepackage{amsthm}
\usepackage{booktabs}
\usepackage{tabularx}
\usepackage[backend=bibtex, maxbibnames=10, style=alphabetic]{biblatex}
\newcommand\citet[1]{\citeauthor*{#1}~\cite{#1}}

\usepackage{color}
\usepackage{amsfonts,amssymb,bbm,bm}
\usepackage{empheq}
\usepackage{cleveref}
\usepackage{xcolor}
\usepackage{tcolorbox}
\usepackage{tikz}
\usetikzlibrary{shadows,arrows.meta,positioning,backgrounds,fit,chains,scopes}
\usepackage{caption}
\usepackage{subcaption}

\usepackage[noend,ruled]{algorithm2e}

\SetCommentSty{mycommfont}
\SetKwInput{KwInput}{Input}
\SetKwInput{KwOutput}{Output}  
\SetAlFnt{\small}
\SetAlCapFnt{\small}
\SetAlCapNameFnt{\small}
\SetAlCapHSkip{0pt}
\IncMargin{-\parindent}

\newcommand{\argmin}{\mathrm{argmin}}

\newcommand{\set}[1]{\left\{ #1 \right\}}

\renewcommand{\tilde}{\widetilde}

\newcommand{\calL}{\mathcal{L}}

\newcommand{\calD}{\mathcal{D}}
\newcommand{\calT}{\mathcal{T}}
\newcommand{\bx}{\bm{x}}
\newcount\Comments  
\newcommand{\kibitz}[2]{\ifnum\Comments=1{\color{#1}{#2}}\fi}

\newcommand{\norm}[1]{\left\lVert #1 \right \rVert}
\title{Meta-Learning with Graph Neural Networks: Methods and Applications}

\author{\ Debmalya Mandal \\
Max Planck Institute for Software Systems \\
Saarbr\"{u}cken, Germany \\
\texttt{dmandal@mpi-sws.org}\\
\And
Sourav Medya \\
Kellogg School of Management\\ Northwestern University, Illinois\\
\texttt{sourav.medya@kellogg.northwestern.edu}\\
\And 
Brian Uzzi\\
Kellogg School of Management\\ Northwestern University, Illinois\\
\texttt{uzzi@kellogg.northwestern.edu} \\
\And
Charu Aggarwal\\
IBM T. J. Watson Research Center\\
Yorktown Heights, New York\\
\texttt{charu@us.ibm.com}
}

\begin{document}

\maketitle
\begin{abstract}
    Graph Neural Networks (GNNs), a generalization of deep neural networks on graph data have been widely used in various domains, ranging from drug discovery to recommender systems. However, GNNs on such applications are limited when there are few available samples. Meta-learning has been an important framework to address the lack of samples in machine learning, and in recent years, the researchers have started to apply meta-learning to GNNs. In this work, we provide a comprehensive survey of different meta-learning approaches involving GNNs on various graph problems  showing the power of using these two approaches together. We categorize the literature based on proposed architectures, shared representations, and applications. Finally, we discuss several exciting future research directions and open problems.
\end{abstract}

\input{1_intro}
\input{2_background_gnn}
\input{3_background_meta}

\input{5_meta_app}

\input{4_meta_and_gnn}

\input{6_future}

\section{Conclusion}
In  this  survey,  we have performed a comprehensive review of the works that are combination of graph neural networks (GNNs) and meta-learning. Besides outlining backgrounds on GNNs and meta-learning, we have organized the past research in an organized manner in multiple categories. We have also provided a thorough review, summary of methods, and applications in these categories. Furthermore, we have described several future research directions where meta learning with GNN can be useful. The application of meta-learning to GNNs is a growing and exciting field and we believe many graph problems will benefit immensely from the combination of the two approaches.
\printbibliography
\end{document}

%% file: 1_intro.tex
\section{Introduction}

The methods of artificial intelligence (AI) and machine learning have found tremendous success in various applications, ranging from natural language processing \cite{Bert18} to cancer screening \cite{WPPS+19}. Such success of AI systems can be attributed to various architectural innovations, and the ability of deep neural networks (DNN) to extract meaningful representations from Euclidean data (e.g. image, video etc.). However, in many applications, the data is graph-structured. For example, in drug discovery, the goal is to predict whether a given molecule is a potential candidate for a new drug, where the input molecules are represented by graphs. In a recommender system, the interaction between the users and the items are represented by a graph, and such non-Euclidean data is crucial in designing a better system.

The proliferation of graph structured data in various applications has led to Graph Neural Networks (GNNs) which are generalizations of DNN for graph-structured inputs. The main goal of GNNs is to learn effective representations of the graphs. Such representations map the vertices, edges, and/or graphs to a low-dimensional space, so that the structural relationships in the graph are reflected by the geometric relationships in the representations~\cite{HYL17}. In recent years, GNNs have been applied in diverse domains, often with surprising positive results like discovery of a new antibiotic~\cite{SYSJ+20}, accurate traffic forecasting~\cite{CHKW19}, etc. 

Despite of recent success of GNNs in various domains, GNN frameworks have their own shortcomings. One of the major challenges in applying GNNs, particularly for large graph-structured datasets, is the limited number of samples. Furthermore, real-world systems like recommender systems often need to handle diverse types of problems, and must adapt to a new problem with very few observations. In recent years, meta-learning has turned out be an important framework to address these shortcomings of deep learning systems.
The main idea behind meta-learning is to design learning algorithms that can leverage prior learning experience to adapt to a new problem quickly, and learn a useful algorithm with few samples. Such approaches have been quite successful in diverse applications like natural language processing \cite{LHCG19}, robotics \cite{NKLK20}, and healthcare \cite{ZTDZ+19}.


Recently, several meta learning methods to train GNNs have been proposed for various applications. There are two main challenges in applying meta-learning to graph-structured data. First, an important challenge is to determine the type of representation that is shared across different tasks. As GNNs are used for a wide range of tasks from node classification to graph classification, the learned shared representation needs to consider the type of tasks to be solved and this makes the choice and design of architecture quite important for meta-learning. 
Second, in a multi-task setting, we usually have few samples from each task. Thus, the support and query examples have often limited overlap in terms of similarity. For example, in node classification tasks, the nodes rarely are similar in the support and query set of a given task. On the other hand, in link prediction, the support and query edges are often located far away from each other in the graph. Therefore, a major challenge in applying meta-learning to GNNs is to model the dependencies among nodes (or edges) that are far apart (both distance-wise and similarity-wise) from each other in the graph. In this survey, we review the growing literature on meta learning with GNNs. There are several thorough individual surveys on GNNs \cite{ZCZY+18,WPCL+20} and meta-learning \cite{HAMS20}, but we believe this survey is the first effort to categorize and comprehensively review the existing papers on meta learning with GNNs.



\subsection{Our Contributions} Besides providing background on meta-learning and architectures based on GNNs individually, our major contributions can be summarized as follows.
\begin{itemize}
\item \textbf{Comprehensive review:} We provide a comprehensive review of meta learning techniques with GNNs on several graph problems. We categorize the literature based on methods, representations and applications and show various scenarios where limitations of GNNs are addressed via meta learning.
\item \textbf{Future directions:} We discuss how meta learning and GNNs can address some of the challenges in several areas: (i) combinatorial graph problems, (ii) graph mining problems, and (iii) other emerging applications such as traffic flow prediction, molecular property prediction, and network alignment.
\end{itemize}

The rest of this paper is organized  as  follows. Section \ref{sec:background_gnn} provides background on a few key graph neural network architectures. Section \ref{sec:meta_background} outlines the background on meta-learning and major theoretical advances. A comprehensive categorization of the papers that use the framework of meta-learning equipped with GNNs on important graph related problems is described in Sections \ref{sec:meta-gnn-1} and \ref{sec:meta-gnn-2}. 
First,  Section \ref{sec:meta-gnn-1} covers applications of meta-learning framework for solving some classical graph problems. The problem discussed here doesn't explicitly propose a multi-task setting, rather the meta-learning framework is applied to a fixed graph. In Section \ref{sec:meta-gnn-2} we cover the literature on graph meta learning when there are multiple tasks and the graph might change with the tasks. Although various GNNs have been proposed for graph meta-learning, they can be categorized broadly based on the type of shared representation, which can be either at a local level (node/edge based) or at the global level (graph based). Table~\ref{tab:summary-global} provides an overview of various papers categorized by the type of shared representation and the application domains. Table \ref{tab:summary-meta-gnn} presents  the papers described in Section \ref{sec:meta-gnn-2} based on the corresponding meta-learning approaches. 
Section \ref{sec:applications} covers a broad range of applications of meta-learning on GNNs and Section \ref{sec:future} suggests some exciting future directions. 


%% file: 2_background_gnn.tex
\section{Graph Neural Networks}
\label{sec:background_gnn}

Generalizing deep learning on graphs has resulted in an exciting area of Graph Neural networks (GNNs). GNNs embed or represent nodes as points in a vector space with the help of structural and attribute information from the neighbourhood of a node and the node itself. They encode this information via non-linear transformations and aggregation functions into a final representation. The proposed architectures can be broadly categorized into two types: (i) \textit{convolution on neighborhood}, and (ii) \textit{location-aware}. 

(i) \textbf{Convolution on neighborhood:} The primary examples of architectures that are based on \textit{convolution on neighborhood} include \textsc{GCN}~\cite{kipf2016semi}, \textsc{GraphSage}~\cite{hamilton2017inductive}, and \textsc{GAT} ~\cite{velivckovic2017graph}. These architectures mostly create representations of nodes through a \textit{convolution} operation $\psi$ over its neighborhood, i.e., the embedding, $z_{v,G}=\psi\left(N^k_{G}(v)\right)$ where the ($k$-hop) neighborhood (set of nodes) of the node $v$ in the graph $G$ is $N^k_{G}(v)$. Thus, two nodes with similar neighborhoods are likely to have similar embeddings. 

(ii) \textbf{Location-aware: } The examples of GNNs that are location aware framework include \textsc{PGNN}~\cite{you2019pgnn} and \textsc{GraphReach} \cite{graphreach}. In this approach, if two nodes are located close (usually by number of hops) to each other in the graph then they are expected to have similar embeddings. If the graph has a high clustering coefficient, then one-hop neighbors of a node share many other neighbors among them as well. Therefore, if two nodes are close to each other, they have a high likelihood of having similar neighborhoods. Many real graphs have \textit{small-world} and \textit{scale-free} properties and have high clustering coefficients. Next, we briefly describe the key architectures of GNNs.

\textbf{GCN \cite{kipf2016semi}: }A primary contribution in applying neural architectures on graphs has been made by \cite{kipf2016semi} with the introduction of Graph Convolutional Networks (GCNs). GCNs are analogous version of convolutional neural networks (CNNs) on graphs. Inspired by the idea of representing a pixel with information from its nearby pixels (filter in CNNs), graph  convolutions  also apply the key idea of aggregating feature information from a node’s local neighborhood. More formally, GCNs are neural network architectures that produces a $d$-dimensional embeddings for each node by taking as input adjacency matrix $A$  and node features $X$; $\mathbf{GCN}(A,X): \mathbb{R}^{n\times n} \times \mathbb{R}^{n \times p} \to \mathbb{R}^{n \times d}$. The idea is to aggregate feature information from a node’s neighborhood (can be generalized to multiple hops) and its own features to produce the final embedding. A 2-layer (neighbourhood is 2-hops) $GCN$ can
be defined as follows:
\begin{equation*}
\mathbf{GCN}(A,X)=\sigma(\hat{A}\sigma(\hat{A}XW^{(1)})W^{(2)})
\label{eqn::gcn}
\end{equation*}
where $\hat{A}=\!\tilde{D}^{-\frac{1}{2}}\tilde{A}\tilde{D}^{-\frac{1}{2}}$ is the normalized adjacency matrix with $\tilde{D}$ as weighted degree matrix and $\tilde{A}\!=\!I_n\!+\!A$ with $I_n$ being an $n\times n$ identity matrix and  $\sigma$ is an activation function. Moreover, $W^{(i)}$ is a weight matrix for the $i$-th layer to be learned during training, with $W^{(1)} \in \mathbb{R}^{p\times d'}$, $W^{(2)} \in \mathbb{R}^{d'\times d}$, and $d$ ($d'$) being the number of neural network nodes in the output (hidden) layer. 

\textbf{\textsc{GraphSage}~\cite{hamilton2017inductive}: } \citet{hamilton2017inductive} propose an inductive framework with an aggregation function that is able to share weight parameters ($\mathbb{W}^k$) across nodes, can be generalized to unseen nodes and scale to large datasets. 
To learn representation $h^{k}_v$ of a node $v$, it iterates over all nodes which are in their K-hop neighborhood. While iterating over node $v$, it \emph{aggregates} (with $\textsc{Aggregate}_k$) the current representations of $v$'s neighbors ($\mathbf{h}^k_N(v)$) and \emph{concatenate} with the current representation of $v$ ($h^{k-1}_v$), which is then fed through a fully connected layer with an activation function. Intuitively, with more iterations, nodes incrementally receive information from neighbors of higher depth (i.e., distance). More specifically for $k$-th iteration,
\begin{equation*}
\begin{split}
\mathbf{h}^k_N(v) &= \textsc{Aggregate}_k \left(\left\{h^{k-1}_u, \forall u\in N(v)\right\}\right) \\
\mathbf{h}^k_v &= \sigma \left (\mathbb{W}^k \cdot \textsc{Concat} \left (\mathbf{h}^k_N(v), h^{k-1}_v\right ) \right )
\end{split}
\end{equation*}

\textbf{\textsc{GAT} \cite{velivckovic2017graph}: }
Graph Attention Networks (GATs) \cite{velivckovic2017graph} learn edge weights using attention mechanisms. GAT does not assume that the contributions of neighbouring nodes are all equal unlike in \textsc{GraphSage} \cite{hamilton2017inductive}.  GAT  learns the relative importance/weights  between  two  connected nodes. The graph convolutional operation ($k$-th iteration) is defined as follows:
\begin{equation*}
\begin{split}
\mathbf{h}^k_v &= \sigma \left (\sum_{u\in N(v)\cup v}\alpha_{v,u}^k\mathbb{W}^k h^{k-1}_v \right )
\end{split}
\end{equation*}
where $\alpha_{v,u}$ measures the strength between the node $v$ and its neighbour $u\in N(v)$. GAT has been shown to outperform both GCN and \textsc{GraphSage} in node classification task both in transductive as well as inductive settings in benchmark datasets.

\textbf{\textsc{PGNN} ~\cite{you2019pgnn}: }
Unlike in \textsc{GraphSage} where the representation of a node depends on its k-hop neighborhood, \textsc{PGNN} follows a different paradigm and aims to incorporate positional information of a node with respect to the nodes in the entire network. The key idea is that the position of a node can be captured via a low-distortion embedding by quantifying the distance between that node and a set of anchor nodes. The framework first samples multiple sets of anchor nodes.  It also learns a non-linear aggregation scheme to combine the features of the nodes in each anchor set. The aggregation is normalized by the distance between the node and the anchor-set.

\textbf{Other variations: }There are  several other variations and improvements of GNNs that are based on different mechanisms: GAT is further extended by Gated Attention Network (GAAN) \cite{zhang2018gaan} through a self-attention mechanism which computes an additional attention score for each attention head. Graph  Autoencoders \cite{cao2016deep,kipf2016variational} encode  nodes/graphs  into  a  latent vector space and further reconstruct the graph related data depending on the application from this encoding in an unsupervised fashion; Recurrent GNNs \cite{scarselli2008graph,li2015gated} apply the same set of parameters recurrently over nodes to extract high-level node representations. For a comprehensive survey on GNNs, please refer to \cite{WPCL+20}.

\subsection{Applications} 

GNNs outperform traditional approaches for semi-supervised learning tasks (e.g. node classification) on graphs. The high level applications of GNNs can be categorized in three major tasks: node classification, link prediction, and graph classification. For node classification and link prediction tasks, traditionally four benchmark datasets are used: Cora, Citeseer, Pubmed, and protein-protein interaction (PPI) dataset. Shchur et al. \cite{shchur2018pitfalls} and Errica et al. \cite{errica2019fair} provide a detailed comparison of performances of the key architectures on node and graph classification tasks. GNNs are also used in the link prediction task that has applications in many domains such as friend or movie recommendation, knowledge graph completion, and metabolic network reconstruction \cite{zhang2018link}.


%% file: 3_background_meta.tex
\section{Background on Meta-Learning}
\label{sec:meta_background}

Meta-learning has turned out to be an important framework to address the problem of limited data in various machine learning applications. The main idea behind meta-learning is to design learning algorithms that can leverage prior learning experience to adapt to a new problem quickly, and learn a useful algorithm with few samples~\cite{Schmid87}. Such approaches have been quite successful in diverse applications like natural language processing \cite{LHCG19}, robotics \cite{NKLK20}, and healthcare \cite{ZTDZ+19}.

\subsection{Framework} In standard supervised learning, we are given a training dataset $\calD = \{\bx_i, y_i\}_{i=1}^n$, a loss function $\ell$, and we aim to find a predictive model of the form $\hat{y} = f_\theta(\bx)$. 
\begin{equation*}
    \theta^* = \argmin_\theta \calL(\calD, \theta) = \argmin_{\theta} \sum_{i=1}^n \ell(f_\theta(\bx_i), y_i)
\end{equation*}
In meta-learning, we are given samples from a number of different tasks and the goal is to learn an algorithm that generalizes across tasks. In particular, the tasks are drawn from a distribution $p(\calT)$, and the meta-objective is to find a common parameter that works across the distribution of tasks.
\begin{equation}\label{eq:meta-obj}
    \omega^* = \argmin_\omega \sum_{\substack{\calT_i \sim p(\calT) \\ \calD_i \sim \calT_i} } \calL_i(\calD_i, \omega)
\end{equation}
In the meta-test phase, we are given a target task (say task $0$) and we use the meta-knowledge $\omega^*$ to obtain the best parameter for the target with few samples.
\begin{equation*}
    \theta_0^* = \argmin_\theta \calL_0(\calD_0, \theta | \omega^*)
\end{equation*}

\subsection{Training} 
Many popular meta-learning algorithms are based on gradient descent on the meta-parameter $\omega$ \cite{FAL17,RL16}. In order to understand how to perform gradient descent with respect to $\omega$, it is insightful to frame Equation \eqref{eq:meta-obj} as a bi-level optimization problem.
\begin{align*}
    \omega^* &= \argmin_\omega \sum_{\substack{\calT_i \sim p(\calT) \\ \calD_i \sim \calT_i} } \calL(\calD_i ,\theta_i^*(\omega), \omega) \\
    \textrm{s.t. } \theta_i^*(\omega) &= \argmin_\theta  \calL_i(\theta,\omega, \calD_i) \ \forall i
\end{align*}
If we have a model for the inner-optimization method, then a gradient of the objective with respect to $\omega$ can be computed by using the chain rule e.g. $$\nabla_\omega \calL(\calD_i ,\theta_i^*(\omega), \omega) = \nabla_{\theta^*_i(\omega)} \calL(\calD_i ,\theta_i^*(\omega), \omega) \frac{d \theta_i^*(\omega)}{d \omega}$$ However, often the inner objective function is non-convex, and hard to solve. So model agnostic meta learning (MAML), introduced by \citet{FAL17} suggests to first take a gradient step for each task $i$ as follows: 
\begin{equation*}
    \theta_i' = \theta_i(\omega) - \alpha \nabla_{\theta} \calL_i(\theta_i(\omega), \omega, \calD_i)
\end{equation*}
Then MAML replaces $\theta_i^*(\omega)$ in the outer objective, i.e.,
\begin{equation*}
    \omega = \omega - \beta \nabla_\omega \sum_i \calL(\calD_i ,\theta_i', \omega)\footnote{We write $\theta_i(\omega)$ to denote the meta-parameter $\omega$ adapted to task $i$.}
\end{equation*}


We now instantiate the MAML algorithm for the task of classifying nodes of a graph. Recall the GCN framework from Section \ref{sec:background_gnn}. Here the $t$-th task is classification of nodes of a graph $G_t$ with adjacency matrix $A_t$ and node-feature matrix $X_t$. Then a standard two-layer GCN for node classification problem is given as follows:
\begin{equation}\label{eq:gcn-ft}
    f(X_t,A_t, W_t) = \textrm{softmax}\left(\hat{A}_t \textrm{ReLU}\left(\hat{A}_t X_t W^{(1)}_t \right)W^{(2)}_t \right)
\end{equation}
Given labels of the nodes $Y_t$, such a network is often trained with the cross-entropy loss:
\begin{equation*}
    \calL_t(X_t, A_t, W_t) = - \sum_{\ell} \sum_{f} Y_{\ell f} \ln f(X_t, A_t, W_t)_{\ell f}
\end{equation*}
Usually, the parameters $W_t$ are trained by stochastic gradient descent. Here, we wish to identify a meta parameter vector $W_\star$, which is close to the parameters of different tasks (i.e. $\norm{W_t - W_{\star}}_F \le \delta$ for some $\delta > 0$). The benefit of learning such meta-parameters $W_\star$ is that, on a new task $s$, we can initialize task-parameter $W_s$ as $W_\star$ and the new task would require very few samples to train. Algorithm~\ref{alg:MAML-GCN} describes the MAML algorithm instantiated for the case of node classification with GCN based representation.

\begin{algorithm}[!ht]
\SetAlgoLined
\DontPrintSemicolon
\KwInput{Step sizes $\alpha$ and $\beta$.}
Initialize $W_{\star}$.\\
\Repeat{Convergence}
{
Sample a batch of $T$ tasks $\{G_i\} \sim p(\cdot)$.\\
Sample a batch of $T$ datasets $\{\calD_i = (A_i, X_i, Y_i)\}$ where $\calD_i\sim G_i$.\\
\For{each task $t$ in $T$}{
Update $W_t = W_\star - \alpha \nabla_{W} \calL_t(X_t, A_t, W)\bigg \rvert_{W=W_\star}$.\\
}
Update $W_\star = W_\star - \beta \nabla_W \sum_t \calL(X_t, A_t, W)\bigg \rvert_{W=W_t}$\\
}
\Return{Meta-parameter $W_\star$.}
\caption{Model Agnostic Meta-Learning for GCN\label{alg:MAML-GCN}}
\end{algorithm}

\subsection{Representation Learning}
Another perspective of meta-learning, which will be particularly important for the context of graph neural networks, is learning a shared representation across different tasks. Here we assume that, given an input $x$, the training data from the $t$-th task is generated as $y_t = f_t \circ h (x) + \eta_t$, where $\eta_t$ is some iid noise. Effectively, the function $h$ maps input $x$ to a shared representation and then a task-specific function $f_t$ is applied to generate the task-specific representation. 

During the meta-training phase, we attempt to learn the shared function $h$. Suppose we are given $T$ datasets $\calD_t = \left\{(x_{ti}, y_{ti}\right\}_{i=1}^{n_t}$ for $t=1,\ldots,T$. Then we solve the following optimization problem to recover $h$.
\begin{equation}
    \label{eq:meta-training-shared}
    \underset{h, \left\{f_t\right\}_{t=1}^T}{\argmin} \sum_{t=1}^T \sum_{i=1}^{n_t} \calL_t\left(y_{ti}, f_t(h(x_{ti})) \right) + \mathcal{R}(h) +  \sum_{t} \mathcal{R}(f_t)
\end{equation}
Here $\mathcal{R}(\cdot)$ is some regularization function, and let $\hat{h}, \left\{ \hat{f}_t\right\}_{t=1}^T$ be its solution.
In general, the optimization problem  defined in Equation \eqref{eq:meta-training-shared} is hard to solve unless we make specific assumption about the types of functions. For example, even if we assume $f_t$ is same across the tasks and in fact an identity function, the problem defined in Equation~\eqref{eq:meta-training-shared} can involve learning a general neural network based shared representation $h$. For the special case of linear models, this problem can be solved efficiently (e.g. by using matrix regression \cite{TJJ20}). In this survey, we focus on gradient based methods for learning the shared representation $h$ in equation \eqref{eq:meta-training-shared}, which has been quite successful in practice. In the meta-test phase, we are given samples from a new task $s$ e.g. $\left\{(x_{si},y_{si}) \right\}_{i=1}^{n_s}$. We substitute $\hat{h}$, the estimate of the common representation function $h$, and learn the new task-specific function $f_s$.
\begin{equation*}
    \hat{f}_s \leftarrow \underset{f_s}{\argmin}  \sum_{i=1}^{n_s} \calL_s\left(y_{si}, f_s(\hat{h}(x_{si})) \right) +  \mathcal{R}(f_s)
\end{equation*}

We now instantiate this framework for the task of classifying nodes of a graph. As before, we use two-layer GCN where the model is defined in Equation \eqref{eq:gcn-ft}. However, we now assume that the first layer is shared across different tasks and only the second layer is trained for a new task. In particular, we assume $W_t = [W^\star; W_t^{(2)}]$. Although, the optimization problem in Equation \eqref{eq:meta-training-shared} is NP-hard to solve with this particular type of representation, we can write down an algorithm to solve for the meta-parameter $W^\star$ using gradient descent. Algorithm \ref{alg:Rep-GCN} describes this algorithm and returns the meta-parameter $W^\star$.

\begin{algorithm}[!ht]
\SetAlgoLined
\DontPrintSemicolon
\KwInput{Step sizes $\alpha$ and $\beta$, datasets $\calD_t = \left\{(x_{ti}, y_{ti}\right\}_{i=1}^{n_t}$ for $t=1,\ldots,T$.}
Initialize $W^{\star}$.\\
Initialize $W^{(2)}_t$ for $t=1,\ldots,T$.\\
Set $W_t = [W^\star, W^{(2)}_t]$.\\
\Repeat{Convergence}
{
\For{each task $t$ in $[T]$}{
Update $W_t^{(2)} = W^{(2)}_t - \alpha \nabla_{W} \calL_t\left(\calD_t, [W^\star; W]\right)\bigg \rvert_{W=W^{(2)}_t}$.\\
}
Update $W^\star = W^\star - \beta \nabla_W \sum_t \calL\left(\calD_t, [W;W^{(2)}_t]\right)\bigg \rvert_{W=W^\star}$
}
\Return{Meta-parameter $W^\star$.}
\caption{Shared Representation Learning for GCN\label{alg:Rep-GCN}}
\end{algorithm}

\subsection{Theory}
Despite immense success, we are yet to fully understand the theoretical foundations of meta-learning algorithms. \citet{Baxter00} first prove generalization bound for multitask learning problem, by considering a model where tasks with shared representation are sampled from a generative model. \citet{PM13}, and \citet{MPR16} develop general uniform-convergence based framework to analyze multitask representation learning. However, they assume oracle access to a global empirical risk minimizer. 
 Recently, there have been promising attempts to understand meta learning from representation learning. The main idea is that the tasks share a common shared representation and a task-specific representation \cite{TJJ20B,TJJ20,DHKL+20}. If the shared representation is learned from the training tasks, then the task-specific representation for the new task can be learned with only a few samples.
 Finally, there have been interesting recent work trying to understand gradient-based meta-learning. \cite{FRKL19,BKT19,KBT19,DCGP19} analyze gradient based meta-learning in the framework of online convex optimization (OCO). They assume that the parameters of the tasks are close to a shared parameter to bound regret in the OCO framework.

%% file: 5_meta_app.tex
\begin{table*}[ht]
    \centering
    \begin{tabular}{lrrr}
    \toprule
    & \multicolumn{3}{c}{Graph applications} \\
    \cmidrule(lr){2-4}
    Representation & Node classification & Link Prediction & Graph Classification \\
    \midrule
    Node/Edge Level  &  Meta-GNN~\cite{ZCZT+19}, GPN~\cite{DWLS+20},  & MetaR~\cite{CZZC+19}  & \\
     &  RALE~\cite{LFLH21}, AMM-GNN \cite{WLDZ+20}   & GEN~\cite{BLH20} &  \\
      & SAME \cite{BV20} SELAR~\cite{hwang2021self} & SAME~\cite{BV20}  & SAME~\cite{BV20}\\
      & GFL~\cite{YZWJ+20}, Meta-GDN~\cite{ding_www21} & SELAR~\cite{hwang2021self} & \\
      \hline
    Graph Level  &MI-GNN~\cite{wen2021meta} & Meta-graph~\cite{BJMH19} & AS-MAML~\cite{MBYZ+20} \\ & & & Spectral~\cite{CNK20} \\
    \bottomrule
    \end{tabular}
    \caption{Organization of the papers on Meta-learning and GNNs based on applications and underlying graph-related representations. The abbreviations of the frameworks (methods) are as follows. GPN: Graph Prototypical Networks, MetaR: Meta Relational learning, GEN: Graph Extrapolation Networks, RALE: Relative and Absolute Location Embedding, AMM-GNN: Attribute Matching Meta-learning Graph Neural Networks, SAME: Single-task
Adaptation for Multi-task Embeddings, SELAR: SELf-supervised Auxiliary LeaRning, GFL: Graph Few-shot Learning, GDN: Graph Deviation Networks, MI-GNN: Meta-Inductive framework for Graph Neural Network, AS-MAML: Adaptive Step Model Agnostic Meta Learning. 
}
    \label{tab:summary-global}
\end{table*}

\section{Meta-Learning on Fixed Graphs}\label{sec:meta-gnn-1}
In this section, we review applications of meta-learning for solving some classical problems on graphs. Here we consider the setting when the underlying graph is fixed and the node/edge features do not change with different tasks. In fact, we are not in a multitask framework where there are a number of tasks and few samples are available from each task. Rather, the framework of  meta-learning is applied to various graph problems by creating multiple tasks either considering the nodes or the edges.

\subsection{Node Embedding} The goal of node embedding is to learn representations for the nodes in the graph so that any downstream application can directly work with these representations, without considering the original graph. This problem is often challenging in practice because the degree distributions of most graphs follow a power law distribution and there are many nodes with very few connections. 
\citet{LZFZH20} address this issue by applying meta-learning to the problem of node embedding of graphs. They set up a regression problem with a common prior to learn the node embeddings. Since the base representations of high-degree nodes are accurate, they are used as meta training set to learn the common prior. The low degree nodes have only a few neighbors (samples), the regression problem for learning their representations is formulated as a meta-testing problem, and the common prior is adapted with a small number of samples for learning the embeddings of such nodes.

\subsection{Node Classification} The node classification task aims to infer the missing labels of nodes of a given partially labeled graph. This problem often appears in diverse contexts such as document categorization and protein classification \cite{TZYL+08,BOSV+05}, and has received significant attention in recent years. However, often many classes are novel i.e., they have a small number of labeled nodes. This makes meta-learning or few-shot learning particularly suitable for this problem.

\citet{ZCZT+19} have applied a meta-learning framework for the node classification problem on graphs by learning a transferable representation using data from classes that have many labeled examples. Then, during the meta-test phase, this shared representation is used to make predictions for novel classes with few labeled samples. 
\citet{DWLS+20} improve upon the previous method by considering a prototype representation of each class and meta-learning the prototype representation as an average of weighted representations of each class.
\citet{lan2020node} address the same  problem via meta-learning but in a different setting where the nodes do not have attributes. Their method only uses the graph structure to obtain latent representation of nodes for the task. 
Subsequently, \citet{LFLH21} point out that it is important to also learn the dependencies among the nodes in a task, and propose to use nodes with high centrality scores (or hub nodes) to update the representations learned by a GNN. This is done by selecting a small set of hub nodes and for each node $v$, considering all the paths to the node $v$ from the set of hub nodes. It helps to encode the absolute location in the graph. Parallel to these developments, \citet{YZWJ+20} consider a metric-learning based approach where the label of a node is predicted to be the nearest class-prototype in a transferable metric space. They first learn a class-specific representation using a GNN, and then learn a task-specific representation using  hierarchical graph representations.

 Finally, the few-shot node classification task has also been used in the presence of noisy or inaccurate labels in the support sets of different tasks. \citet{ding2021weakly} propose a method (Graph Hallucination Network) that creates a set by taking a specified number of samples from a class. Then the method learns to produce a confidence score on the accuracy of the label of each node in the set. By using these weights/scores, the final cleaner (i.e., less noisy) node representations are generated. The rest of the algorithm follows the standard MAML framework. 


\subsection{Link Prediction} 
The objective of the link prediction problem is to identify pairs of nodes that will either form a link or not. 
Meta-learning has been shown to be useful for learning new relationship via edges/links especially in multi-relational graphs. 

In multi-relational graphs, an edge is represented by a triple of two end points and a relation. Such graphs appear in many important domains such as drug-drug interaction prediction. The goal of link prediction in multi-relation graphs is to predict new triples given one end point of a relation $r$ with observing a few triples about $r$. This problem is challenging as only few associative triples are usually available. \citet{CZZC+19} use meta-learning to solve the link prediction problem in two steps. First, they design a Relation-Meta Learner which learns shared structure across a number of relations. Such a meta-learner
generates relation meta from heads' and tails' embeddings in the support set. Second, they use an embedding learner that calculates the truth values of triples in support set via end points' embeddings and relation meta. 

Multi-relational graphs are even more difficult to manage with their dynamic nature (addition of new nodes) over time and the learning is even more difficult when these newly evolved nodes have only few links among them. \citet{BLH20} introduce a few-shot out-of-graph link prediction technique, where they predict the links between the seen and unseen nodes as well as between the unseen nodes. The main idea is to randomly split the entities in a given graph into the meta-training set for simulated unseen entities, and the meta-test set for real unseen entities.

Finally, \citet{hwang2021self} show the effectiveness of graph neural networks on downstream tasks such as node classification and link prediction via a self-supervised auxiliary learning framework combined with meta-learning. 
The auxiliary task such as meta-path prediction does not need labels and thus the method becomes self-supervised. In the meta learning framework, various auxiliary tasks are used to improve generalization performance of the underlying primary task (e.g., link prediction). The proposed method effectively combines the auxiliary tasks and automatically balances them to improve performance on the primary task. The method is also flexible to work with any graph neural network architecture without additional data.

%% file: 4_meta_and_gnn.tex

\section{Meta-Learning on Graph Neural networks}\label{sec:meta-gnn-2}
We now discuss the growing and exciting literature on graph meta learning where there are multiple tasks and the underlying graph can change across the tasks. The changes in graphs occur when either the node/edge features change, or the underlying network structure changes with the tasks. In the context of meta-learning, several architectures have been proposed in recent years. However, a common thread underlying all of them is a shared representation of the graph, either at a local node/edge level, or at a global graph level. Based on the type of shared representation, we categorize the existing works into two groups. Most of the existing literature adopt the MAML algorithm \cite{FAL17} to train the proposed GNNs. The outer loop of MAML updates the shared parameter, whereas the inner loop updates the task-specific parameter for the current task. Table \ref{tab:summary-meta-gnn} lists the shared and the task-specific parameters for all the papers in this section.

  \begin{table}
\small
\vspace{2mm}
    \centering
    \begin{tabular}{ccc}
    \toprule
    & \multicolumn{2}{c}{Meta-learning parameters} \\
    \cmidrule(lr){2-3}
    Papers & Inner Loop & Outer Loop \\
    & (Task-Specific) & (Shared)\\
    \hline
    \cite{HZ20} & Node embeddings & Classification\\
    \hline
    \cite{WLDZ+20} & Node embeddings & Feature matrix\\
    \hline
    \cite{CNK20} & Graph feature, & graph label/ \\
     & Super-class & actual class \\
     \hline 
     \cite{MBYZ+20} & Graph feature, & graph embedding/\\
     & Graph embedding & Classification\\
     \hline
     \cite{BV20} & Node Embedding & Output Layer\\
     \hline
     \cite{BJMH19} & VGAE Initialization & Graph Signature\\
     & & (GCN + MLP) \\
     \hline 
     \cite{LZFZH20} & High-degree & node specific  \\
     & node embedding &  embedding \\
   
    
    \bottomrule
    \end{tabular}
    \caption{Organization of the papers in Section 6 based on the corresponding meta-learning approaches.}
    \label{tab:summary-meta-gnn}
\end{table}





\begin{figure*}[!h]
    \centering
    \begin{minipage}[!b]{0.48\linewidth}
    \includegraphics[width=\linewidth]{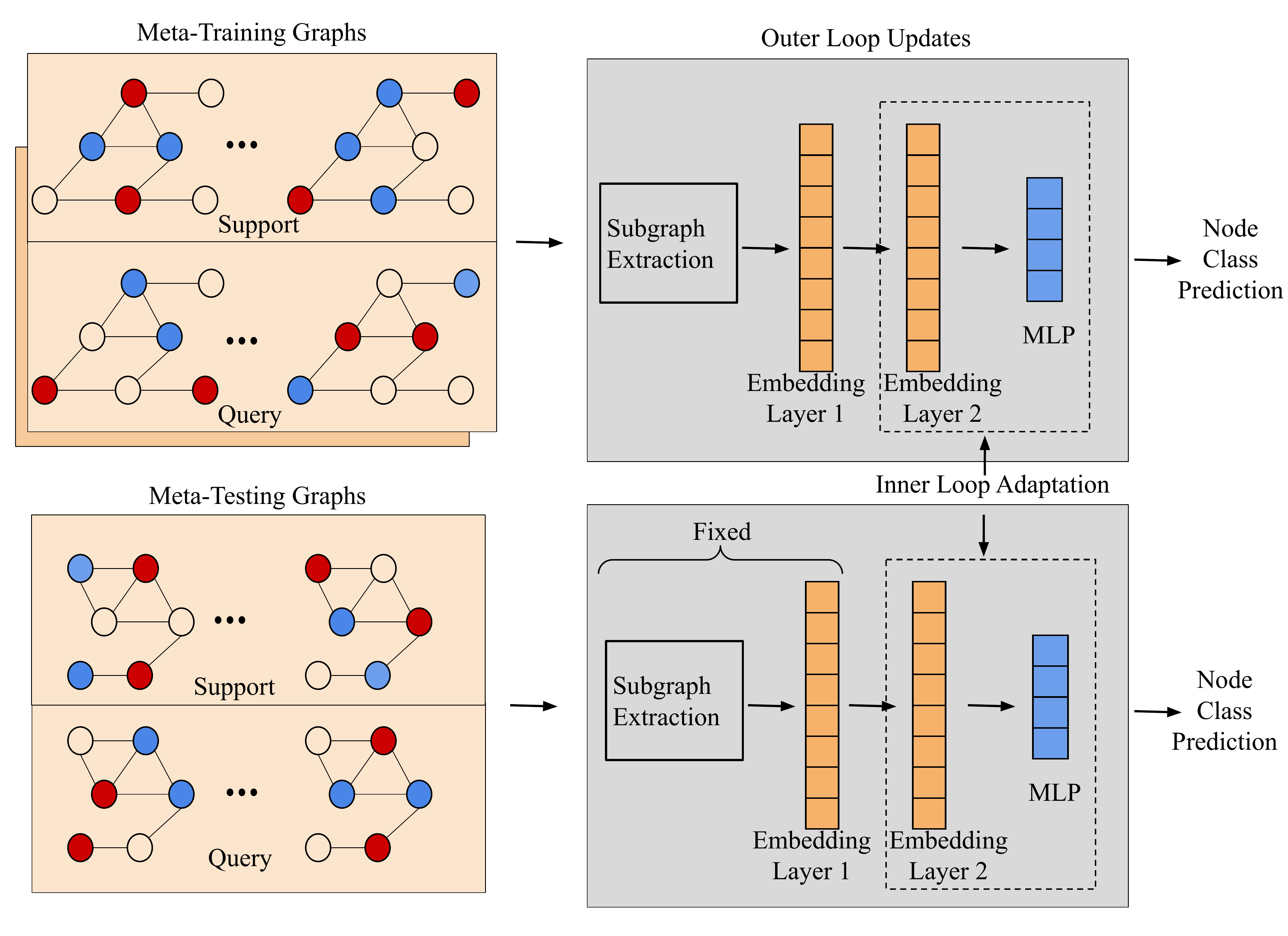}
    \caption{A prototype of the meta learning framework with GNNs for solving node classification problem. This is based on the architectures proposed by \protect\cite{HZ20} and \protect\cite{WLDZ+20}. Following \protect\cite{HZ20}, the neighborhoods of each node are used for node embedding. Embedding layer 1 is trained in the outer loop of MAML, whereas the other layers are adapted for particular tasks.}
    \label{fig:meta-gnn-local}
    \end{minipage}\hspace{0.01\linewidth}
    \begin{minipage}[!b]{0.48\linewidth}
     \includegraphics[width=\linewidth]{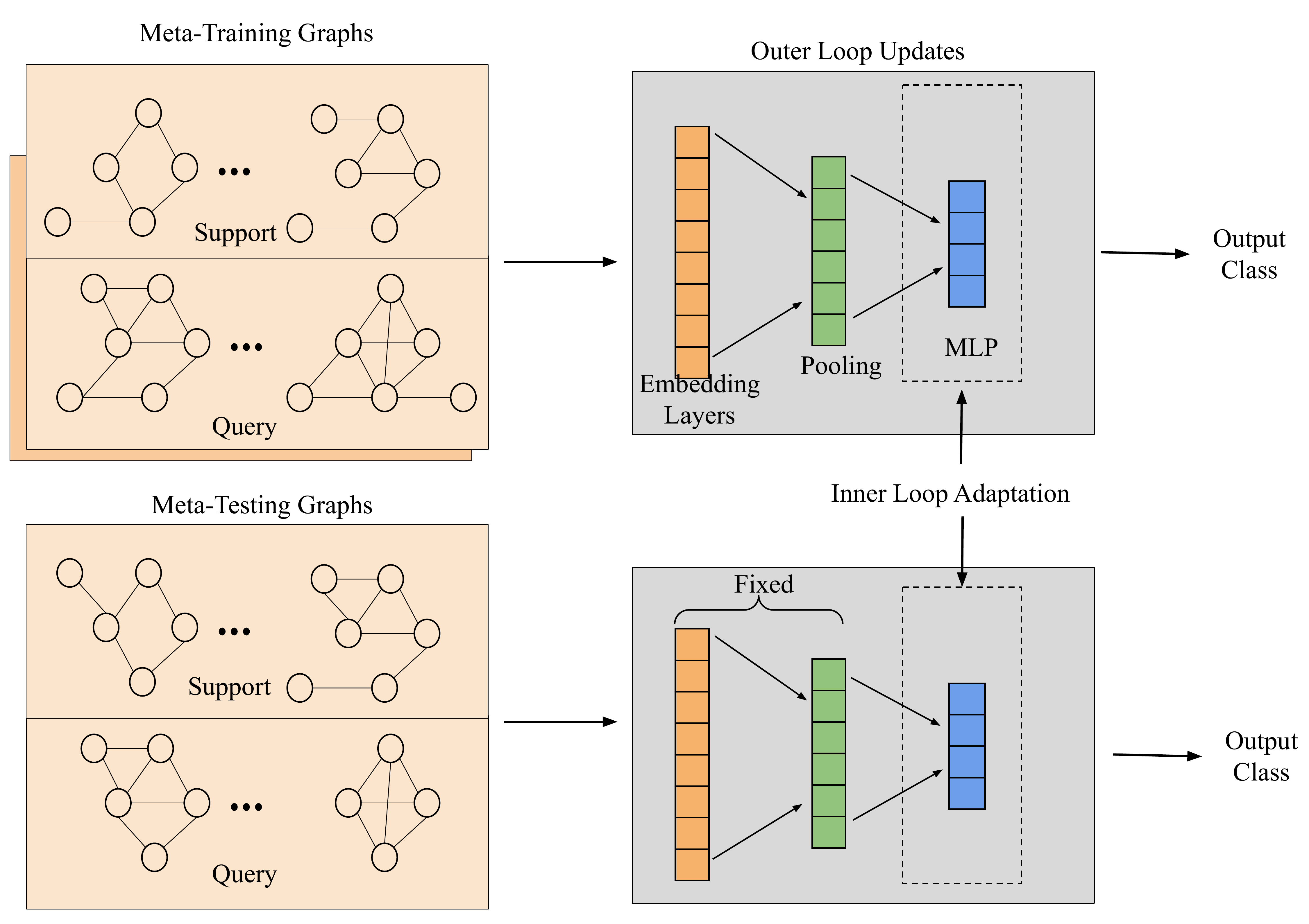}
    \caption{A prototype of the meta learning framework with GNNs for solving graph classification problem. This is based on the architectures proposed by \protect\cite{MBYZ+20}, and \protect\cite{BV20}. The embedding and pooling layers learn global representation of the input graph, and are trained in the outer loop of MAML. The final multi-layer perceptron (MLP) is used for the classification task and is adapted to the particular task at meta-test.}
    \label{fig:meta-gnn-global}
    \end{minipage}
    \vspace{1mm}
\end{figure*}
\subsection{Node/Edge Level Shared Representation}
First, we consider the setting where the shared representation is local i.e. node or edge based. \citet{HZ20} consider the node classification problem where the input graphs as well as the labels can be different across tasks. They learn a representation for each node $u$ in two steps. First, the method extracts a subgraph $S_u$ corresponding to the set of nodes $\set{v: d(u,v) \le h}$ where $d(u,v)$ is the distance of the shortest path between nodes $u$ and $v$. Then it feeds the subgraph $S_u$ through a GCN to learn a representation for node $u$. The theoretical motivation behind considering the graph $S_u$ is  
that the influence of a node $v$ on $u$ decreases exponentially as the shortest-path distance between them increases. Once the nodes are encoded, one can learn any function $f_\theta$ that maps the encodings to class labels. \citet{HZ20} use MAML to learn this function with very few samples on a new task, enjoying the benefits of node-level shared representations in node classification.

\citet{WLDZ+20} also consider the few-shot node classification problem for a setting where the network structure is fixed, but the features of the nodes change with tasks. In particular, given a base graph with node feature matrix $X \in \mathbb{R}^{n \times d}$, the proposed model learns a new feature matrix $X_t = X \odot \alpha_t(\phi) + \beta_t(\phi)$ for the $t$-th task, and then use a GNN $f_\theta(X_t)$ to learn the node representations for the $t$-th task. During training, the outer loop updates the $\phi$ parameters, whereas the inner loop of MAML only updates the $\theta$-parameter. This enables quick adaptation to the new task.

\citet{wen2021meta} study the problem of node classification in an inductive setting, where the graph instances in  testing and training do not overlap. Their method involves computing a task prior given a graph (i.e., its representation) using multi-layer perceptron (MLP). These representations are useful for the graph-level adaptation. They used the traditional MAML paradigm in their approach for the task-level adaptation.
 

\subsection{Graph Level Shared Representation}
In this subsection, we discuss the setting when the shared representation is global i.e. graph-level. A canonical application of this representation is the \emph{graph classification} problem, where the goal is to classify a given graph to one of many possible classes. This problem appears in many applications, ranging from bioinformatics to social network analysis~\cite{YV15}. However, in many settings, the number of samples/graphs available for a particular task is few and the graph classification task often requires a large number of samples for high quality prediction. 
These challenges can be addressed via meta-learning. The existing papers on using meta-learning for graph classification usually learn an underlying shared representation and adapt the representation for a new task.

\citet{CNK20} propose the few-shot graph classification task based on graph spectral measures. In particular, they train a feature-extractor $F_\theta(\cdot)$ to extract features from the graphs in meta-training. For classification, they first use a unit $C^\text{sup}$ to first predict the super-class probability of a graph which is a clustering of abundant base class labels. Then they use $C^\text{att}$, an attention network to predict the actual class label. During the meta-test phase, the weights of the networks $F_\theta(\cdot)$ and $C^\text{sup}$ are fixed, and the network $C^\text{att}$ is retrained on the new test classes. As the feature extractor $F_\theta$ is the common shared structure, and is not retrained on the test tasks, this approach requires few samples from new classes.

Although \citet{CNK20} propose a novel meta-learning architecture for graph classification, there are several limitations. First, the architecture assumes significant overlap between the super-class structure of the test and the training set. Second, the fixed feature extractor cannot be updated for the new tasks. \citet{MBYZ+20} design a better meta-learning technique by allowing the feature extractor to adapt efficiently for new tasks. They apply two networks -- embedding layers ($\theta_e$), followed by classification layers ($\theta_c$) to classify a given graph. However, for a new task, both $\theta_e$ and $\theta_c$ are updated. In particular, the authors use MAML \cite{FAL17} to update the parameters and use a reinforcement learning based controller to determine how the inner loop is run i.e., what is the optimal adaptation step for a new task. The parameters of the controller is updated using the graph's embedding quality and the meta-learner's training state.

\citet{jiang2021structure} solve the problem of few-shot graph classification via a paradigm in meta learning called metric learning approach \cite{wang2019simpleshot} that is different from MAML. In the training phase, the idea is to get a mean representations of the instances in each class in the support set. The prediction for query is based on the nearest neighbour. Here the graph representations were obtained by the Graph Isomorphism Network (GIN) model. To capture the global structure of the graph, they used different weights for different GIN layers in the final aggregation scheme. To encode the crucial local structures that might have importance in deciding the graph label, the paper embeds subgraphs and includes their representations with different attention weights. 

Finally, \citet{BV20} attempt to develop a framework that can adapt to three different tasks -- graph classification, node classification, and link prediction. Like \cite{CNK20,MBYZ+20} they use two different layers; one generates node embeddings and converts the graph to a representation, and another is a multi-head output layer for the three types of tasks. The node embedding layer is trained during the initialization phase of MAML and the multi-head output layer is updated in the inner loop of MAML based on the type of task.

\citet{BJMH19} consider the few-shot link prediction problem, where the goal is to predict labels of links/edges that contain only a small fraction of their true labels. They assume that the graphs are generated from a common distribution $p(\cdot)$ and learn a meta link prediction model that can be quickly adapted to a new graph $G \sim p(\cdot)$. In particular, the authors use Variational Graph Autoencoder (VGAE) \cite{kipf2016variational} to model the base link prediction model. There are two sets of parameters -- global initialization parameters for the VGAE, and local graph signature $s_G = \psi(G)$ which is obtained by passing the graph $G$ through GCN and then using a $k$-layer MLP. The training is done using MAML and only the graph signature is updated for the test graph.

%% file: 6_future.tex
\section{Other Applications}
\label{sec:applications}
We have discussed applications of meta-learning equipped with GNNs on node classification, link prediction, and graph classification. In fact, this framework is quite general and can be applied to many other relevant important problems.


\textbf{Anomaly Detection:} The problem of anomaly detection often suffers from scarcity of labels, as obtaining labels for anomalies is usually labor intensive. 
\citet{ding_www21} study anomaly detection when there are scarcity of labels, and  different tasks involve different graphs. The proposed method used traditional architectures of GNNs to embed nodes and predict the anomaly score by adding another layer after the embedding is obtained. Finally it  exploits the traditional MAML framework to deploy the meta-learner. The inner loop optimizes the parameters for a specific task, i.e., graph. The outer-loop optimizes the generic parameter for all graphs.


\textbf{Network Alignment (NA): } NA aims to map or link entities from different networks and relevant in many application domains such as cross-domain recommendation and advertising. \citet{zhou2020fast} address this alignment problem via meta-learning. If two different networks share some common nodes or anchors, then these  networks are partially  aligned  networks. A virtual link between two anchors is called anchor link. In NA, given a set of networks and some known anchor nodes (or links), the goal is to identify all the other (unknown) potential anchor nodes (or links). The main idea in \cite{zhou2020fast} is to frame this problem as one shot classification problem and use the meta-metric  learning from  known anchor nodes to obtain latent priors for linking  unknown  anchor nodes.

\textbf{Traffic Prediction: }Recently, the traffic prediction  problem \cite{pan2020spatio} has been addressed via meta-learning. In traffic prediction, the main challenges are modeling complex spatio-temporal correlations of traffic and capturing the diversity of such correlations varying locations. 
\citet{pan2020spatio} address these challenges with a meta-learning based model. Their method predicts traffic in all locations at the same time. The proposed framework consists of a sequence-to-sequence architecture that uses an an encoder to learn traffic history and a decoder to make predictions. For the encoder and decoder components a combination of graph attention networks and recurrent neural networks is used to model diverse spatial and temporal correlations respectively.


\section{Future Directions}
\label{sec:future}

The application of meta-learning using GNNs for graph specific applications is a growing and exciting area of research. In this section, we suggest several future directions for research.

\subsection{Combinatorial Optimization Problems on Graphs}

Combinatorial optimization problems appearing in graphs have applications in many domains such as viral marketing in social networks \cite{kempe2003maximizing}, health-care \cite{Wilder2018aamas}, and infrastructure development~\cite{medya2018noticeable}, and several architectures based on GNNs have been proposed for solving them \cite{dai2017learning,li2018combinatorial,gasse2019exact,manchanda2020gcomb}.
 These optimization problems are often NP-hard, and polynomial-time algorithms, with or without approximation guarantees, are often desirable and used in practice. However, some techniques \cite{li2018combinatorial,manchanda2020gcomb} based on GNNs need to generate candidate solution nodes/edges before generating the actual solution set. Note that, labels in the form of importance of each node in a solution set of these problems are often difficult to get. Meta-learning can be used when there are scarcity of labels. Furthermore, these combinatorial problems often share similar structures. For instance, the influence maximization problem \cite{kempe2003maximizing} have similarity with the Max Cover problem. However, even performing a greedy iterative algorithm to generate solutions/labels for influence maximization problem is computationally expensive. The idea of using meta-learning in solving a harder combinatorial problem (unseen task) with a fewer node labels will be to learn on the easier problems (seen tasks) where labels can be generated at a lower cost. Solving combinatorial optimization problems on graphs via neural approaches has recently gained a lot of attention and we refer the readers to \cite{cappart2021combinatorial} for further reading. 
 \subsection{Graph Mining Problems} 
 There has been recent attempt to solve classical graph mining problems with GNNs. For instance, a popular problem is to learn similarity between two graphs, i.e., to find graph edit distance (similarity) between two graphs \cite{BDBCSW19}. When the notion of similarity changes and there are not enough data to learn via a standard supervised learning method, can meta-learning be helpful? Another popular graph mining problem is detecting the Maximum Common Subgraph (MCS) between two input graphs with applications in biomedical analysis and malware detection. In drug design, common substructures in compounds can reduce the number of human-conducted experiments. However, MCS computation is NP-hard, and state-of-the-art exact MCS solvers are not scalable to large graphs. Designing learning based models \cite{bai2020fast} for the MCS problem while utilizing as few labeled MCS instances as possible remains to be a challenging task and meta-learning could be helpful in mitigating this challenge. \\

\subsection{Theory}
We point out several important theoretical questions in the context of meta learning with GNNs. The most natural question is understanding the benefits of transfer learning in GNNs. \citet{GJJ20} and \citet{STH18} have recently established generalization bounds for GNNs. On the other hand, in the context of meta-learning, \citet{TJJ20B}  consider functions of the form $f_j \cdot h$, where $f_j \in \mathcal{F}$ is the task-specific function and $h$ is the shared function. Then the number of samples required in the meta-test phase grows as $C(\mathcal{F})$, which can be significantly lower than learning $f_j \cdot h$ from scratch. It would be interesting to see if one can prove similar speedup results for GNNs by generalizing the results of \cite{GJJ20} and \cite{STH18}. Another interesting question is determining the right level of shared representation and figuring out the expressiveness of such structures. The seminal work of \citet{XHLJ18} proves that variants of GNNs such as GCN and GraphSAGE are no more discriminative than the Weisfeiler-Lehman (WL) test. Since GNNs for meta-learning further limit the type of architecture used, an interesting question is whether it comes with any additional cost on expressiveness. Finally, the methods discussed in Section \ref{sec:meta-gnn-2} differ in one crucial way -- whether they fine-tune and update the shared meta-parameter on a new task or whether they keep the shared meta-parameter fixed. Recently, \citet{KLL21} show that fine-tuning the meta-parameter could be beneficial in some situations, particularly when the number of samples on the new task is large. In the context of meta learning on GNNs, it would be interesting to understand when such fine-tuning helps to improve the performance on a new task.

\subsection{Applications}
We have already discussed a few applications of meta-learning with frameworks of GNNs in Section \ref{sec:applications}. This generic framework is quite relevant for many important problems in the field. 


\textbf{Network alignment: }A potential problem where meta-learning could be helpful is network alignment (NA) \cite{zhou2020fast}. In NA, the main goal is to map or link entities from different networks and the existing approaches is quite difficult to scale. An interesting direction of research would consider meta-learning to overcome this scalability challenge.

\textbf{Molecular property prediction: }GNNs have been also used in predicting molecular properties. However, one of the main challenges is that molecules are heterogeneous structure where each atom has connection with different neighboring atoms via different types of bonds. Secondly, often a limited amount of data on labeled molecular property are available; and thus, to predict new molecular properties, meta-learning techniques \cite{guo2021few} can be relevant and effective.  

\textbf{Dynamic graphs:}
In many applications, graphs arise with their dynamic nature, i.e., nodes and edges along with their attributes can change (addition or deletion) over time. Most of the papers discussed above use frameworks that are built on meta-learning and GNNs for static graphs. An interesting direction would be to extend this framework for dynamic graphs. Dynamic nature brings new challenges such as difficulty in obtaining labels for newly added nodes or edges. For instance, in knowledge graphs, newly added edges introduces new relationships. The other challenge is efficiency as managing and making predictions on evolving networks are difficult tasks as its own. Meta-learning would be useful to address these challenges.

